\def\year{2022}\relax
\documentclass[letterpaper,english]{article} 
\usepackage{aaai22}  
\usepackage{times}  
\usepackage{helvet}  
\usepackage{courier}  
\usepackage[hyphens]{url}  
\usepackage{graphicx} 
\usepackage{natbib}  
\usepackage{caption} 
\DeclareCaptionStyle{ruled}{labelfont=normalfont,labelsep=colon,strut=off} 
\usepackage{algorithm}
\usepackage{algorithmic}
\usepackage{newfloat}
\usepackage{listings}
\floatstyle{ruled}
\newfloat{listing}{tb}{lst}{}
\floatname{listing}{Listing}

\usepackage{color}
\usepackage{multirow}
\usepackage{amsmath}
\usepackage{amsfonts}

\title{LEA-Net: Layer-wise External Attention Network \\ for Efficient Color Anomaly Detection}
\author {
    Ryoya Katafuchi,\textsuperscript{\rm 1}
    Terumasa Tokunaga, \textsuperscript{\rm 1}
}
\affiliations {
	\textsuperscript{\rm 1} Kyushu Institute of Technology\\ 
    katafuchi.ryoya556@mail.kyutech.jp, tokunaga@ai.kyutech.ac.jp
}

\begin{document}
\maketitle

\section{Abstract}

	The utilization of prior knowledge about anomalies is an essential issue for anomaly detections. 
	Recently, the visual attention mechanism has become a promising way to improve the performance of CNNs for some computer vision tasks. 
	In this paper, we propose a novel model called Layer-wise External Attention Network (LEA-Net) for efficient image anomaly detection. 
	The core idea relies on the integration of unsupervised and supervised anomaly detectors via the visual attention mechanism. 
	Our strategy is as follows: 
		(i) Prior knowledge about anomalies is represented as the anomaly map generated by unsupervised learning of normal instances, 
		(ii) The anomaly map is translated to an attention map by the external network, 
		(iii) The attention map is then incorporated into intermediate layers of the anomaly detection network. 
	Notably, this layer-wise external attention can be applied to any CNN model in an end-to-end training manner. 
	For a pilot study, we validate LEA-Net on color anomaly detection tasks. 
	Through extensive experiments on PlantVillage, MVTec AD, and Cloud datasets, 
		we demonstrate that the proposed layer-wise visual attention mechanism consistently boosts anomaly detection performances of an existing CNN model, even on imbalanced datasets. 
	Moreover, we show that our attention mechanism successfully boosts the performance of several CNN models.

\section{1. Introduction}
	\label{sec:introduction}
	
	\begin{figure}[ht]
		\centering
		\captionsetup{justification=raggedright, singlelinecheck=off}
		\includegraphics[width=\hsize]{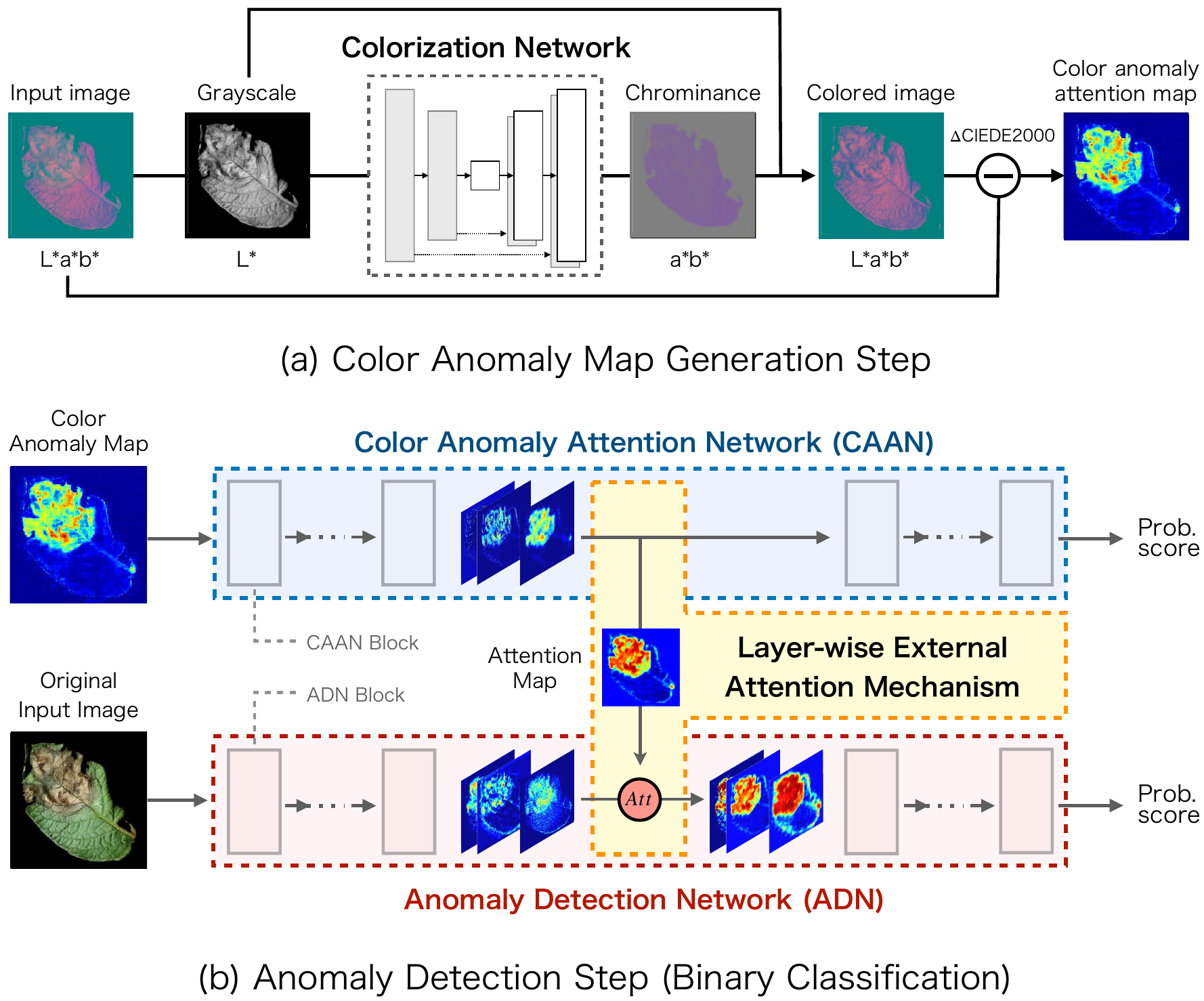}
		\caption{Overview of the color anomaly detection.} 
		\label{fig:Overview}
	\end{figure}

	Anomaly detection is a technique to identify irregular or unusual patterns in datasets.
	Especially, anomaly detection for imaging data is a powerful and core 
	technology for various kinds of real-world problems including medical diagnosis~\cite{rezvantalab2018dermatologist,cao2018deep}, 
	plant healthcare~\cite{Ferentinos_2018}, 
	production quality controls, and disaster detection~\cite{Minhas,8995282}. Over the last decade, many researchers have shown
	great interest in establishment of automatic anomaly detection techniques for a huge image dataset 
	driven by breakthroughs in deep learning. Based on types of machine learning, these anomaly detection techniques can be roughly 
	classified into three categories: supervised, semi-supervised, and unsupervised
	approaches. Although there are advantages and disadvantages with each approach, the fundamental 
	challenge that should be overcome is on how we detect anomalies efficiently based on limited numbers of anomalous instances.
		
	Convolutional neural network (CNN) is a commonly used artificial network for various computer vision tasks including image recognition and image segmentation.
	Using a huge dataset of labeled images, CNNs have realized a state-of-the-art image anomaly detection in a real-world applications~\cite{Hughes_2016, Minhas}. 
	However, anomaly detectors based on CNN suffer from a lack of labeled images and low anomaly instances.
	Some studies have proposed ideas for overcoming this point by improving learning efficiency of CNN. In particular, one introduced active learning~\cite{goernitz2013} and the other employed transfer learning~\cite{Minhas} for that purpose. 
	
	Approaches based on unsupervised learning are the most popular way for anomaly detection, because they do not require labeled anomalous instances to train anomaly detectors.
	The simple strategy of unsupervised image anomaly detections relies on a training of reconstruction processes for normal images using a deep convolutional
	autoencoder~\cite{haselmann2018anomaly}. However, the autoencoder sometimes fails to reconstruct fine structures. Consequently, it outputs immoderate blurry structures. 
	Recently, generative adversarial networks (GAN) have been used for image anomaly detection to overcome this point. AnoGAN~\cite{Schlegl2017} firstly employed
	GAN to image anomaly detection.  In applications of medical image processing,  AnoGAN and its extensions~\cite{Zenati_2018, Akcay_2018} realized to detect 
	minute anomalies. More recently, the idea of AnoGAN was extended in terms of color reconstructability to realize sensitive detection of color anomalies~\cite{Katafuchi2021}. 
	It is a common procedure in unsupervised anomaly detection to define the difference between the original image and the reconstructed image 
	as {\it Anomaly Score}.

	Although unsupervised anomaly detectors can eliminate the labeling cost of anomalous instances for a training step, 
	there are some drawbacks. First, they tend to overlook small and minute anomalies because {\it Anomaly Score}
	is defined based on the distance of the normal images and test images.  In other words, the detection performance 
	strongly depends on whether we can design a good {\it Anomaly Score} for each purpose. Also, we should tune carefully 
	the appropriate threshold of {\it Anomaly Score} for successful classifications of normal and anomalous instances. In most practices, 
	this process requires painful trial and error. 	
	
	Nowadays, the technique called visual attention mechanism is attracting much attention in computer vision~\cite{Zhao2020}.
	Attention Branch Network (ABN) is a CNN with a branching structure called {\it Attention Branch}~\cite{Fukui}. Attention map that output from 
	{\it Attention Branch} represent visual explanations, which enables humans to understand the decision-making process of CNN. Further, the authors
	reported that the attention map contributes to improving the performances of CNN for several image classification tasks. Since then,  
	researchers in computer vision have attempted to build the technique that finely induces the attention of CNN to informative regions in images. 
	For the last several years, 
	it has been shown that such technique is promising for improving the performance of CNN for several computer vision tasks involving image classification 
	and segmentation~\cite{Fukui,emami2020spa}. Inspired by these studies, one may think to introduce the visual attention mechanism to image
	anomaly detection. However, visual attention modules including ABN basically relies on self-attention mechanisms~\cite{Hu2020,Woo2018}. 
	Accordingly, the attention quality of the modules strongly depends on the performance of the network itself, which means that existing visual attention 
	mechanisms would not be able to boost the performance of anomaly detectors directly.

	In this paper, we propose Layer-wise External Attention Network (LEA-Net) to boost CNN-based anomaly detectors.
	As we mentioned above, purely unsupervised anomaly detectors do not utilize anomalous instances at all for training.
	As a result,  they tend to overlook small and minute anomalies, or conversely, incur high false positives. 
	On the other hand, purely supervised approaches suffer from lack of labeled images, especially from lack of anomalous instances. 
	We tackle these problems by integrating supervised and unsupervised anomaly detectors via visual attention mechanism. 
	Recent progress in visual attention mechanism strongly implies that there must be a way to utilize prior knowledge for anomaly detection. 
	Our expectation is that the anomaly map that generated from an unsupervised anomaly detector is powerless by itself for efficient anomaly 
	detection but useful to boost supervised anomaly detectors. 
	Moreover, we expect that such a boosting approach contributes to reduce developing cost of anomaly detectors because we can divert 
	existing image classifiers for image anomaly detection. 

	Our overall strategy is as follows: (i) Prior knowledge on anomalies is represented as the anomaly map, which is generated through unsupervised learning of normal instances, 
	(ii) The anomaly map is translated to an attention map by the external network, (iii) The attention map is then incorporated into intermediate layers of the anomaly 
	detection network. We note that this layer-wise external attention can be applied to any CNN model with an end-to-end training manner. 
	For a pilot study, we focus on color anomaly detection tasks because color anomalies are comparatively easy to represent based on CIEDE2000 color difference~\cite{Katafuchi2021}. 
	We examined the effectiveness of LEA-Net through extensive experiments using real-world publicly available datasets including PlantVillage, MVTec AD, and Cloud datasets. The results demonstrated that the proposed layer-wise visual attention mechanisms consistently boost performances of anomaly detectors even on imbalanced datasets. 
	Moreover, it was shown that the attention mechanism works well for multiple feed-forward CNN models.
	
	Our contributions are as follows:
	\begin{itemize}
		\item We propose a novel anomaly detection model called Layer-wise External Attention Networks (LEA-Net) for efficient image anomaly detection
		with a limited number of anomalous instances.
		\item We showed that our layer-wise external attention mechanism successfully boosts the performance of the color image anomaly detectors, even for small 
		and imbalanced training datasets.
		\item We found that the layer-wise external attention mechanism is effective for existing feed-forward CNN models.
		\item We provide our code and models as open source at https://github.com/RxstydnR/LEA-Net. 
	\end{itemize}

\section{2. Related Work}
	The proposed approach is an integration of unsupervised image translation and supervised anomaly detection.
	Most recently, a simpler approach was adopted in a task of automatic identification for thyroid nodule lesions in X-ray computed tomography images~\cite{Li2021}.
	The technique uses binary segmentation results obtained from U-Net as inputs for supervised image classifiers. 
	The authors showed that the binary segmentation as a preprocessing for an image classification contributes to improving anomaly detection
	in a real-world problem. Looking for a somewhat similar setting, Convolutional Adversarial Variational autoencoder with Guided Attention (CAVGA) 
	uses an anomaly map in a weakly supervised setting to localize anomalous regions~\cite{venkataramanan2020attention}. Through experiments for 
	image anomaly detection using MVTec AD, CAVGA achieved SOTA results. 
	These two studies imply that the introduction of a visual attention map has a great potential for image anomaly detection. 
		
	A visual attention mechanism typically refers to the process of  refinement or enhancement of image features for recognition tasks.
	The human perceptual system tends to preferentially capture information relevant to the current task, rather than processing all information~\cite{reynolds2004attentional,chun2011taxonomy}. 
	Visual attention mechanisms basically imitate this for image classification~\cite{Hu2020,Wang2017,Woo2018,Lee2019,Wang2020,yang2021simam}.
	Most image classifiers with visual attention adopt a self-attention module that works in plug-and-play to existing models.
	In such specifications, the effectiveness of visual attention strongly depends on the performance of the main body of the model: this is a limitation of 
	self-attention approaches.  Attention Branch Network (ABN)~\cite{Fukui} overcame this problem by interactive edit of attention map. 
	It enabled us to induce focus points of CNN to more informative regions on images through the correction of attention map. Another similar visual 
	attention mechanism is Attention Transfer based on knowledge distillation~\cite{Zagoruyko2017}. In knowledge distillation, a smaller network called 
	student network receives prior knowledge from a larger network called teacher network~\cite{hinton2015distilling}.
	The idea of Attention Transfer relies on an assumption that a teacher network tends to focus on a more informative area than does a student network.
		
	We designed LEA-Net inspired by ABN and Attention Transfer Network with several points of improvement.
	Our method does not require any user-interaction to control focus points of CNN. The training of LEA-Net goes on 
	fully automatically through a collaboration of two networks that we call external network and anomaly detection network.
	The external network adjusts the strength of attention with progress of training to avoid excessive effects of visual attention 
	mechanism in the early stage of training. Also, we note that the effectiveness of the layer-wise external attention mechanism 
	is not constrained by the teacher network unlike Attention Transfer Network.

\section{3. Proposed Method}

	\begin{figure*}[t]
		\centering
		\captionsetup{justification=centering, labelfont=bf}
		\includegraphics[width=0.8\hsize]{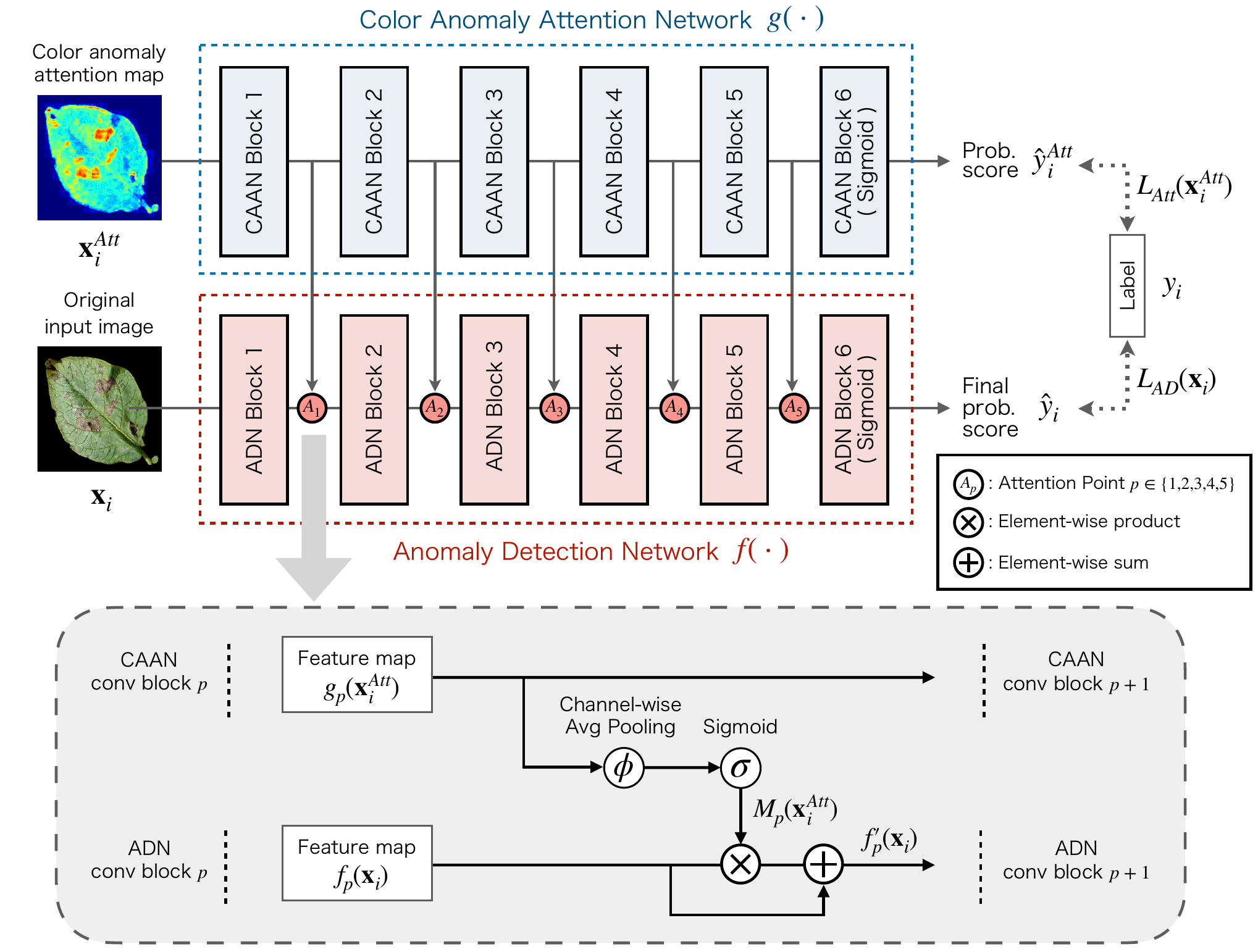}
		\caption{Overall structure of LEA-Net.}
		\label{fig:AttentionMechanism}
	\end{figure*}

	As we mentioned in Section~1, we focus on color anomaly detection tasks for a pilot study.
	In Fig.\ref{fig:Overview}, we illustrate the overview of our color anomaly detection. The procedure
	can be divided into two steps: (a) color anomaly map generation and (b) color anomaly detection.
	Details will be described below.

	\subsection{3-1. Color Anomaly Map Generation}
		In step~(a), a grayscale image is firstly obtained from the input color image represented in $L^{*}a^{*}b^{*}$ color space.
		Then, the grayscale image is reconverted to a color image using U-Net~\cite{ronneberger2015u}.
		To supplement, the color information $a^{*}b^{*}$(chrominance) is predicted based on $L^{*}$ (luminance) information in the $L^{*}a^{*}b^{*}$ color space. 
		Second, the predicted $a^{*}b^{*}$ is combined with the $L^{*}$ of input to produce the resulting colored image in $L^{*}a^{*}b^{*}$ color space.
		The U-Net learns in advance the process of color reconstruction with normal instances only. 
		Then, a color anomaly map is generated by calculating CIEDE2000~\cite{Sharma_2005} color difference between the reconstructed image and the original image. 
		For details of the color anomaly generation, see~\cite{Katafuchi2021}. 
		
		Here we consider data augmentation because the concern of this study is on anomaly detection with comparatively few training data. 
		Empirically, the color reconstruction often fails for quite bright or dark images. Hence, we use Fancy PCA image 
		augmentation~\cite{krizhevsky2012imagenet} to amplify the variation of image luminances.

	\subsection{3-2. Color Anomaly Attention Network~(CAAN)}
		We describe the detail of Fig.~\ref{fig:Overview}(b): Anomaly detection step. 
		In this step, we use two different image classifiers. 
		One is Color Anomaly Attention Network~(CAAN), which converts the anomaly map to an attention map. Except for color anomaly detection, we call the network {\it External Network}.
		The other is Anomaly Detection Network~(ADN), which outputs the final decision for anomaly detection.
		
		It is desirable that CAAN is a lightweight and effective model to work in plug-and-play without a massive increase in parameters. 
		To that purpose, architectures of CAAN were designed based on MobileNet and ResNet. Detailed structures of CAAN are described in Table~\ref{tab:CAAN}.
		The structure of MobileNet-based CAAN is exactly same as MobileNetV3-Small \cite{howard2019searching}.
		Besides, the  structure of ResNet-based CAAN is configured by simply stacking five residual blocks~\cite{he2016deep}.
		We note that these two models are generic for image classifiers. Therefore, they can be replaced easily by other networks.
			
	\subsection{3-3. Anomaly Detection Network~(ADN)}					
		The role of ADN is to make the final decision for image anomaly detection utilizing prior knowledge received from CAAN.
		We adopted ResNet18 and VGG16 for ADN, both of which are well-known and widely-used CNN for image classifications.
		The number of downsampling points in ADN should be the same or larger than that of CAAN due to the reasons described below. 
		
		\begin{table}[t]
			\captionsetup{labelfont=bf}
			\caption{Structures of CAAN: MobileNet-based network and ResNet-based network. 
			The convolution layers are denoted as $\{$conv2d, $\langle$receptive field$\rangle$x$\langle$receptive field$\rangle$, $\langle$number of channels$\rangle$$\}$.
			''bneck'' denotes bottleneck structure; see \cite{howard2019searching}. 
			}
			\label{tab:CAAN}
			\centering
			\scalebox{0.8}[0.8]{ 
				
				\renewcommand{\arraystretch}{1.5}
				\begin{tabular}{c|c|c}
					\hline 
						block name & 
						\begin{tabular}{c}
							ResNet-based \\
							CAAN 
						\end{tabular}& 
						\begin{tabular}{c}
							MobileNet-like \\
							CAAN 
						\end{tabular}

					\\ \hline \hline 
						\multicolumn{3}{c} { input (256$\times$256$\times$1 Anomaly map) } 
					\\ \hline 
						block 1 &
						conv2d, 7x7, 64 & 
						conv2d, 3x3, 16 
					
					\\ \hline 
						\multicolumn{3}{c} { Attention Output Point 1 }

					\\ \hline 
						block 2 & 
						\begin{tabular}{l} 
							conv2d, 3x3, 64 \\[-5pt]
							conv2d, 3x3, 64
						\end{tabular} &
						bneck, 3x3, 16 

					\\ \hline 
						\multicolumn{3}{c} { Attention Output Point 2 } 

					\\ \hline 
						block 3 & 
						\begin{tabular}{l} 
							conv2d, 3x3, 128 \\[-5pt]
							conv2d, 3x3, 128
						\end{tabular} & 
						[bneck, 3x3, 24] $\times$ 2 

					\\ \hline 
						\multicolumn{3}{c} { Attention Output Point 3 } 

					\\ \hline 
						block 4 & 
						\begin{tabular}{l} 
							conv2d, 3x3, 256 \\[-5pt]
							conv2d, 3x3, 256
						\end{tabular} & 
						\begin{tabular}{l} 
							[bneck, 5x5, 40] $\times$ 3 \\[-5pt]
							{[bneck, 5x5, 48]} $\times$ 2 
						\end{tabular} 
					
					\\ \hline 
						\multicolumn{3}{c} { Attention Output Point 4 } 
						
					\\ \hline 
						block 5 & 
						\begin{tabular}{l} 
							conv2d, 3x3, 512 \\[-5pt]
							conv2d, 3x3, 512
						\end{tabular} & 
						[bneck, 5x5, 96]$\times$3 
						
					\\ \hline 
						\multicolumn{3}{c} { Attention Output Point 5 } 
						
					\\ \hline 
					
					block 6 & 
						\begin{tabular}{c} 
							average pool \\[-5pt]
							2-d fc \\[-6pt]
							sigmoid
						\end{tabular} & 
						\begin{tabular}{c}
							conv2d, 1x1, 576 \\ [-5pt]
							average pool \\[-5pt]
							conv2d, 1x1, 1280 \\[-5pt]
							conv2d, 1x1, 2 \\[-5pt]
							sigmoid
						\end{tabular}

					\\ \hline 
						Params & 
						4.491M & 
						3.042M   
					\\ \hline
				\end{tabular}
				\renewcommand{\arraystretch}{1}
			}
		\end{table}
		
	\subsection{3-4. The overall structure of LEA-Net}	
	
		In Fig.~\ref{fig:AttentionMechanism}, we illustrate the overall structure of LEA-Net.
		As shown in Fig.~\ref{fig:AttentionMechanism}, CAAN has five feature extraction blocks. In other words, CAAN has five alternatives for outputting attention 
		maps for layer-wise visual attention. Simultaneously, ADN can have attention points up to five. 
		In this study, the attention point in ADN is limited to one to avoid the performance deterioration of ADN.

		In the training process, both CAAN and ADN are optimized by passing through the gradients of the CAAN and ADN network during back propagation.
		Let us denote the $i$th original input image by ${\mathbf x}_{i} \in \mathbb{R}^{H \times W \times 3}$.
		Let ${\mathbf x}^{Att}_{i} \in \mathbb{R}^{H \times W}$ be the $i$th color anomaly map. Also, let $y_{i}\in \left \{0, 1\right \} $ be a corresponding ground-truth label.
		Further, let $L_{Att}$ and $L_{AD}$ be loss functions for CAAN and for ADN, respectively. The loss function for the entire classification network 
		can be expressed as a sum of the two loss functions as:
		\begin{eqnarray}
			\label{eq:Loss}
			L &=& L_{Att}+L_{AD} \nonumber \\ 
				&=& BCE(g({\mathbf{x}_i^{Att}}),y_{i})+BCE(f({\mathbf{x}_i}),y_{i}).
		\end{eqnarray}
		Here, $g(\cdot)$ and $f(\cdot)$ denote the output of CAAN and that of ADN, respectively. Also, $BCE(\cdot)$ denotes the binary crossentropy.
		We designed the loss function expecting that CAAN modifies attention maps more effectively during training likewise ABN.
	
	\subsection{3-5. Attention Mechanism}
	
		Let us denote the $p$-th attention point in ADN by \\ $A_p~(p \in \{1,2,3,4,5\})$. Also, let $M_p~\in~\mathbb{R}^{H_p \times W_p}$
		be the attention map at $A_p$.
		The concept of layer-wise external attention is illustrated in the lower half of Fig.~\ref{fig:AttentionMechanism}.
		Let $g_{p}(\mathbf{x}^{Att}_i) \in \mathbb{R}^{H_p \times W_p \times C_{p}}$ be the feature tensor at attention point $p$ in CAAN for input
		image $\mathbf{x}^{Att}_i$.
		Then, $M_p$ is generated by
		\begin{equation}
			\label{eq:AttentionMap}
			M_p = \sigma(\phi(g_p(\mathbf{x}^{Att}_i))),
		\end{equation}
		where $\phi(\cdot)$ denotes channel-wise average pooling on the extracted features. Also, $\sigma(\cdot)$ denotes a sigmoid function. By a channel-wise 
		average pooling on the extracted features, we obtain the single-channel feature map $\phi_(g_{p}(\mathbf{x}^{Att}_i)) \in \mathbb{R}^{H_p\times W_p}$.
		We adopted channel-wise average pooling rather than a $1 \times 1$ convolution layer, expecting the effect reported in~\cite{Woo2018}.
		A sigmoid layer $\sigma(\cdot)$ normalizes the feature map $\phi(g_{p}(\mathbf{x}^{Att}_i))$ within a range of $[0, 1]$.
		It was reported that the normalization of the attention map is effective to highlight informative regions effectively~\cite{Wang2017}.
		In addition, the sigmoid function prevents ADN features after attention from the reversal of importance by multiplying negative values.
		Then, we obtain the attention map $M_{p}(\mathbf{x}_{i}^{Att}) \in \mathbb{R}^{H_p \times W_p}$.

		The role of the attention mechanism is to highlight the informative regions on feature maps, rather than erasing other regions~\cite{Wang2017}.
		To reduce the risk that an informative region is erased by attention maps, we incorporate the attention map into ADN as follows:
		\begin{equation}
			\label{eq:AttentionMechanism}
			\hat{f_p}(\mathbf{x}_{i}) = (1\oplus M_p) \otimes f_p(\mathbf{x}_{i}), 
		\end{equation}
		where $\oplus$ denotes element-wise sum, $\otimes$ denotes element-wise product, and $\hat{f_{p}}(\cdot)$ is the updated feature tensor at point $p$ in 
		ADN after the layer-wise external attention. 
		
		Our attention strategy described in Eq.~\ref{eq:AttentionMechanism} is also intended to avoid the Dying Relu Problem~\cite{Lu_2020}.
		The problem is that many parameters with negative values become zero when they pass through the Relu function, which will cause the vanishing gradient problem.
		In most cases, CAAN and ADN have significantly different feature maps.
		In addition, as the layers get deeper, the feature maps typically tend to be sparser.
		If such sparse features are simply multiplied at attention points,  the performance of ADN will degrade seriously.
		This is the other reason why we adopt the attention strategy in Eq.~\ref{eq:AttentionMechanism} rather than simply multiplying the attention map.

\section{4. Experiment}
	
	In this section, we evaluate the performance of LEA-Net using several datasets for image anomaly detections. 
	Our experiments involve three main parts.  
	In the first part, we verify the effect of the layer-wise external attention for boosting the performance of existing image classifiers
	using four real-world datasets.  In the second part, we examine the performance of LEA-Net on more imbalanced settings.
	Finally, we attempt to visualize feature maps at all attention points before and after the layer-wise external attention to understand intuitively the effects of our attention mechanism.

	\begin{figure*}[t]
		\hspace{-2.5em}
		\centering
		\captionsetup{justification=raggedright, singlelinecheck=off, labelfont=bf}
		\includegraphics[width=0.8\hsize]{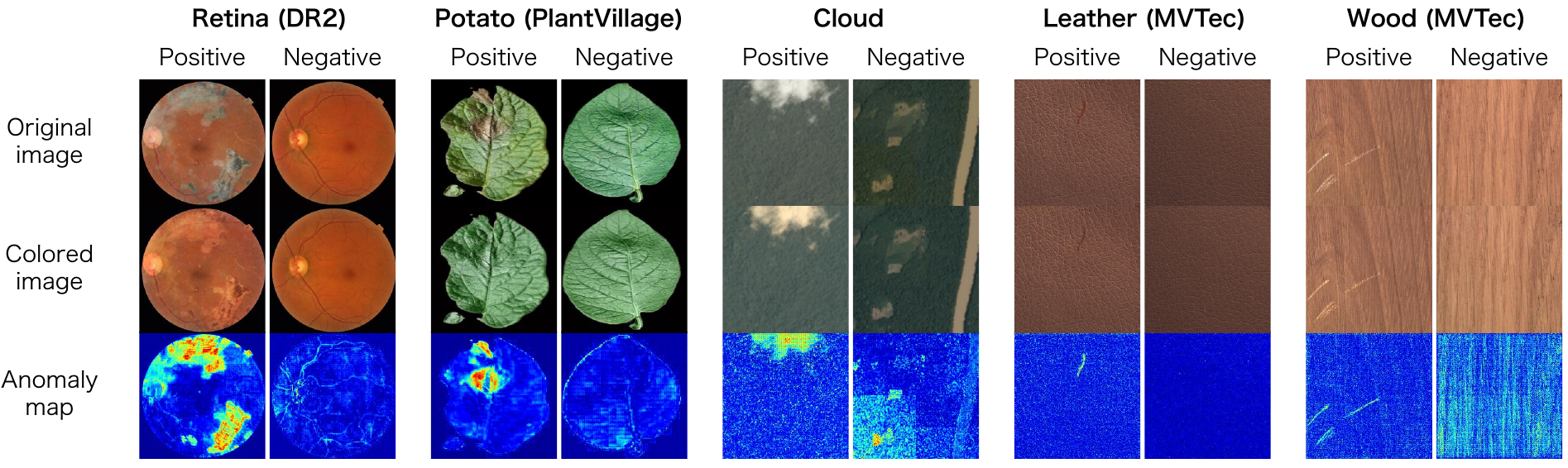}
		\caption{Examples of datasets. 
				 For each dataset, the original and colored image and the anomaly map calculated from the two images are shown for Positive and Negative.
			}
		\label{fig:Dataset}
	\end{figure*}

	\subsection{4-1. Datasets}
		Datasets used here are shown in Fig.~\ref{fig:Dataset}. 
		We performed the experiment for image anomaly detection using DR2 \cite{Pires2016}, PlantVillage \cite{Hughes_2016}, 
		MVTec \cite{Bergmann_2019}, and Cloud \cite{cloud} datasets. 
		DR2 contains $435$ publicly available retina images with sizes of $857 \time 569$ pixels. It consists of normal (negative) and
		Diabetic Retinopathy (positive) instances. PlantVillage contains images of healthy and diseased leaves of several plants.
		Among them, we used \textit{Potato} dataset here. MVTec AD contains defect-free and anomalous images of various objects and texture categories.
		From this dataset, we used \textit{Leather} and \textit{Wood}, whose anomalies are strongly reflected in color.
		Cloud dataset contains images without clouds (negative) and with clouds (positive).
		The lowest panels in Fig.~\ref{fig:Dataset} represent anomaly maps. We see that anomalous regions in the Positive images were failed to reconstruct and 
		highlighted in the color anomaly maps. All images were resized to $256 \times 256$ pixels before experiments.
		To evaluate the performance on such practical dataset, we randomly extracted images from each dataset to construct imbalanced datasets.
		The details of datasets are shown in Table~\ref{tab:Dataset}.

		\begin{table}[t]
			\captionsetup{labelfont=bf}
			\caption{Imbalanced experimental datasets reconstructed by random sampling.}
			\label{tab:Dataset}
			\scalebox{1}[1]{ 
			\renewcommand{\arraystretch}{1.2}
			\hspace{-1.0em}
			\begin{tabular}{c|c c c c}
				
				Dataset Name 
				& Positive 
				& Negative 
				& \begin{tabular}{c}
					P to N \\[-0.5em] Ratio
				\end{tabular} 
				& Total \\ \hline
		
				Retina \scriptsize{(DR2)}     & 98  & 337 & 0.291 & 435  \\
				Potato \scriptsize{(PlantVillage)}    & 50  & 152 & 0.329 & 202  \\
				Cloud      & 100 & 300 & 0.333 & 400 \\
				Leather \scriptsize{(MVTec)}    & 92  & 277 & 0.332 & 369  \\
				Wood \scriptsize{(MVTec)}      & 60  & 266 & 0.226 & 326  \\

			\end{tabular}
			\renewcommand{\arraystretch}{1}
			}
		\end{table}

	\subsection{4-2. Experimental Setup}
		We briefly describe the experimental setup for training.
		Parameters of U-Net were optimized by Adam optimizer with a learning rate of $0.001$. 
		Momentums were set to $\beta_1 = 0.9$ and $\beta_2 = 0.999$. 
		The total number of iterations to update the parameters was $500$ epochs, and the batch size was set to $16$. 
		To reduce the computational time, early stopping was used. 
		
		In the anomaly detection experiment, we performed stratified five-fold cross-validation on each dataset without data augmentation.
		Parameters of ResNet18 were optimized by Adam optimizer with a learning rate of $0.0001$ and 
		those of VGG16 were optimized by Adam optimizer with a learning rate of $0.00001$. 
		The total number of iterations to update the parameters was set to $100$ epochs, and the batch size was $16$. 
		All computations were performed on a GeForce RTX 2028 Ti GPU based on a system running Python $3.6.9$ and CUDA $10.0.130$.

	\subsection{4-3. Performance on Anomaly Detection}
		
		In LEA-Net, the attention map is incorporated into the intermediate layers of ADN via CAAN, which is the main idea of the layer-wise external attention mechanism.
		To evaluate the effectiveness of our attention mechanism, we compared the anomaly detection performance of LEA-Net with several settings:
		(i) baseline models (ResNet18, RestNet50, VGG16 and VGG19),
		(ii) ADN that receives anomaly map as inputs (\textit{Anomaly Map input}), 
		(iii) ADN that receives 4-channel images consist of RGB images and anomaly maps as inputs (\textit{4ch input}), 
		(iv) ADN that receives attention maps generated from RGB images and anomaly maps according to Eq.~\ref{eq:AttentionMechanism} as inputs~(\textit{Attentioned input}),
		(v) ADN with the layer-wise external attention at the attention point without CAAN (\textit{Direct Attention}).
		
		The averages of $F_1$ scores with a standard deviation in each experiment are shown in Table~\ref{tab:Performance}.
		The best and second-best performances are emphasized by bold type. The values given in parentheses indicate attention point of the best $F_1$ score. 
		As mentioned above, the layer-wise external attention was applied at only one point in each experiment although the ADN has five attention points.
		For $F_1$ scores of models with the layer-wise external attention, we described the result of the best attention point score.
		
		The results in Table~\ref{tab:Performance} clearly show that the layer-wise external attention mechanism consistently increased the $F_1$ scores.
		In most of the cases, the layer-wise external attention with CAAN demonstrated better performance compared to that without CAAN. 
		Surprisingly, for \textit{Potato} dataset, the layer-wise external attention increased the $F_1$ score of ResNet18 by $0.304$.
		However, the performances were comparatively low in cases where the anomaly map was directly incorporated into ADN. 
		Interestingly, we must mention here that the most effective attention point disagrees depending on the models and datasets.

		\begin{table*}[t]
			\captionsetup{labelfont=bf}
			\caption{Comparison of $F_1$ scores on Retina, Potato, Cloud, Leather, Wood datasets}
			\label{tab:Performance}
			\scalebox{0.8}[0.8]{ 
			\renewcommand{\arraystretch}{1.5}
			\begin{tabular}{l|c||c|c|c|c|c}
				
				ADN Model   &  
				Params  & 
				Retina  &
				Potato  & 
				Cloud   & 
				Leather & 
				Wood    \\ 
				
				\hline 
				
				\small{ResNet18}           	
									& 11.2M              	
									& 0.642 $\pm$ 0.097  	
									& 0.767 $\pm$ 0.164  	
									& 0.642 $\pm$ 0.081  	
									& 0.756 $\pm$ 0.147  	
									& 0.560 $\pm$ 0.230 \\ 	

				\small{ResNet50}
									& 23.6M 
									& 0.544 $\pm$ 0.155 
									& 0.425 $\pm$ 0.315 
									& 0.635 $\pm$ 0.147 
									& 0.631 $\pm$ 0.152 
									& 0.546 $\pm$ 0.146\\

				\small{ResNet18 (AnomalyMap input)}
									& 11.2M 
									& 0.412 $\pm$ 0.005 
									& 0.779 $\pm$ 0.181 
									& 0.476 $\pm$ 0.058 
									& 0.551 $\pm$ 0.300 
									& 0.329 $\pm$ 0.137 \\

				\small{ResNet18 (4ch input)}
									& 11.2M 
									& 0.443 $\pm$ 0.054 
									& 0.850 $\pm$ 0.102
									& 0.529 $\pm$ 0.050 
									& \textbf{0.883 $\pm$ 0.066} 
									& 0.495 $\pm$ 0.322\\

				\small{ResNet18 (Attentioned input)}
									& 11.2M 
									& 0.616 $\pm$ 0.108 
									& 0.823 $\pm$ 0.157
									& 0.532 $\pm$ 0.050 
									& 0.752 $\pm$ 0.035 
									& 0.544 $\pm$ 0.22\\
				
				\small{ResNet18 + Direct Attention}
									& 11.2M 
									& \textbf{0.718 $\pm$ 0.047(3)}
									& 0.884 $\pm$ 0.045(4)
									& \textbf{0.670 $\pm$ 0.084(3)}
									& 0.823 $\pm$ 0.055(3) 
									& \textbf{0.697 $\pm$ 0.061(4)}\\ 

				\small{\textbf{ResNet18 + MobileNet-small CAAN}}
									& 13.5M 
									& 0.707 $\pm$ 0.037(1)
									& \textbf{0.870 $\pm$ 0.094(4)}
									& \textbf{0.707 $\pm$ 0.131(2)}
									& 0.826 $\pm$ 0.062(1)
									& \textbf{0.719 $\pm$ 0.247(2)} \\ 

				\small{\textbf{ResNet18 + ResNet-based CAAN}}
									& 16.1M 
									& \textbf{0.737 $\pm$ 0.045(1)} 
									& \textbf{0.948 $\pm$ 0.039(4)}
									& 0.664 $\pm$ 0.067(4)
									& \textbf{0.828	$\pm$ 0.052(4)}  
									& 0.621 $\pm$ 0.142(1)\\ 
				\hline
				\small{VGG16}
									& 165.7M              	
									& 0.709 $\pm$ 0.119  	
									& 0.906 $\pm$ 0.061  	
									& 0.575 $\pm$ 0.046  	
									& 0.862 $\pm$ 0.089  	
									& 0.652 $\pm$ 0.142\\ 	

				\small{VGG19}
									& 171.0M 
									& 0.692 $\pm$ 0.068
									& 0.941 $\pm$ 0.04
									& \textbf{0.794 $\pm$ 0.041}
									& 0.758 $\pm$ 0.08
									& 0.597 $\pm$ 0.338\\

				\small{VGG16 (AnomalyMap input)}     
									& 165.7M 
									& 0.399 $\pm$ 0.012
									& 0.899 $\pm$ 0.093
									& 0.438 $\pm$ 0.023
									& \textbf{0.907 $\pm$ 0.039}
									& 0.397 $\pm$ 0.111\\

				\small{VGG16 (4ch input)}
									& 165.7M 
									& 0.458 $\pm$ 0.031
									& 0.823 $\pm$ 0.073
									& 0.479 $\pm$ 0.092
									& \textbf{0.925 $\pm$ 0.033} 
									& 0.575 $\pm$ 0.042\\

				\small{VGG16 (Attentioned input)}        	
									& 165.7M 
									& 0.726 $\pm$ 0.056
									& \textbf{0.947 $\pm$ 0.039}
									& 0.576 $\pm$ 0.119
									& 0.877 $\pm$ 0.053
									& 0.593 $\pm$ 0.269\\
				
				\small{VGG16 + Direct Attention}
									& 165.7M 
									& \textbf{0.754 $\pm$ 0.060(5)}
									& 0.927 $\pm$ 0.030(1)
									& \textbf{0.596 $\pm$ 0.072(4)}
									& 0.877 $\pm$ 0.102(3)
									& \textbf{0.746 $\pm$ 0.069(3)}\\ 

				\small{\textbf{VGG16 + MobileNet-small CAAN}}
									& 168.0M 
									& 0.744 $\pm$ 0.068(3) 
									& 0.938 $\pm$ 0.058(1)
									& 0.577 $\pm$ 0.069(3)
									& 0.867 $\pm$ 0.077(2) 
									& 0.707 $\pm$ 0.074(4)\\ 

				\small{\textbf{VGG16 + ResNet-based CAAN}}
									& 170.6M 
									& \textbf{0.764 $\pm$ 0.069(5)} 
									& \textbf{0.947 $\pm$ 0.064(1)}
									& 0.584 $\pm$ 0.076(5)
									& 0.899 $\pm$ 0.091(3) 
									& \textbf{0.762 $\pm$ 0.035(5)}\\ 
		
			\end{tabular}
			\renewcommand{\arraystretch}{1}
			}
		\end{table*}

	\subsection{4-4. Performance on more imbalanced data}
		We conducted anomaly detection tests using more imbalanced datasets.
		Four datasets from \textit{Retina}~(DR2) and \textit{Potato}~(PlantVillage) were constructed by changing a proportion of the number of anomalous instances to that of all 
		instances. Then, the performance of LEA-Net was compared to that of baseline 
		models (ResNet18 and VGG16). 
		Settings of LEA-Net were as follows: (i) ResNet18-based ADN with MobileNet-small CAAN, (ii) ResNet18-based ADN with ResNet-based CAAN, (iii) VGG16-based ADN with
		MobileNet-small CAAN, (iv) VGG16-based ADN with ResNet-based CAAN.

\begin{figure}[htb]
			\centering
			\captionsetup{justification=raggedright, singlelinecheck=off, labelfont=bf}
			\includegraphics[width=1.0\hsize]{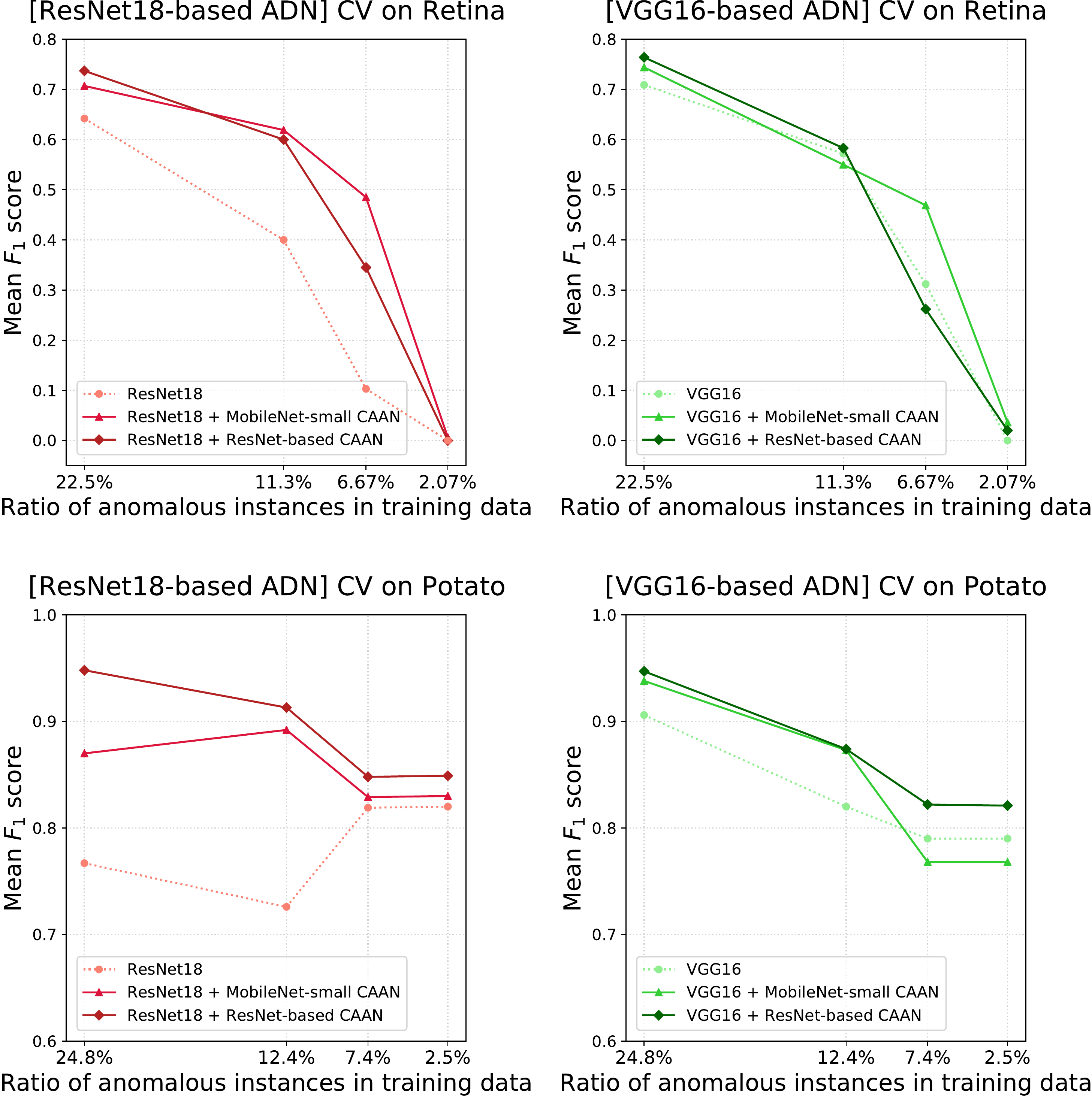}
			\caption{Performances on several imbalanced settings.}
			\label{fig:ImbalancedResult}
		\end{figure}

		The resulting $F_1$ scores for each imbalanced dataset are described in Table~\ref{fig:ImbalancedResult}. 
		The upper two panels show the results for \textit{Retina} and the lower two panels show those for \textit{Potato}.
		All $F_1$ scores were averaged over the five-fold cross-validation. Horizontal axes indicate a proportion of the number of anomalous instances to 
		that of all instances. 
		We see that $F_1$ scores of the ResNet18-based ADN with CAAN were significantly improved by the layer-wise external attention throughout all imbalanced settings. 
		Also, $F_1$ scores of VGG16-based ADN with CAAN for \textit{Potato} were successfully 
		improved in settings of $24.8\%$ and $12.4\%$. 
		However, for $F_1$ scores of VGG16-based ADN with CAAN for \textit{Retina}, there was no meaningful improvement in all settings.

		\begin{figure*}[t]
			\hspace{-1.5em}
			\captionsetup{justification=centering, labelfont=bf}
			\includegraphics[width=1.05\hsize]{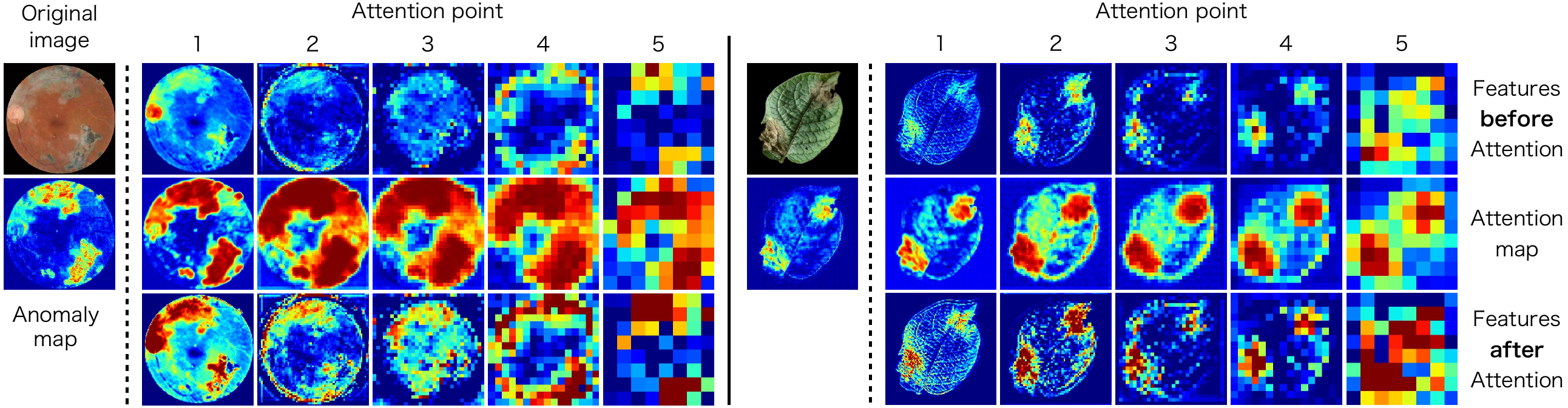}
			\caption{Visualization of feature maps at attention points before and after the layer-wise external attention.}
			\label{fig:Visualization}
		\end{figure*}
	
	\subsection{4-5. Visualization of Attention Effects}
	
		Finally, we attempted to visualize how the layer-wise external attention induces attentions of ADN.
		Fig.\ref{fig:Visualization} displays feature maps output from intermediate layers in ResNet18-based ADN with ResNet-based CAAN
		before and after the layer-wise external attention for DR2 (\textit{Retina}) and PlantVillage (\textit{Potato}) datasets.
		We can see that anomalous regions were substantially highlighted by the layer-wise external attention throughout, from low-level to high-level features.

\section{5. Conclusion}
	
	We proposed a Layer-wise External Attention Network (LEA-Net) for efficient image anomaly detection. 
	LEA-Net has the layer-wise external attention mechanism to introduce prior knowledge about anomalies, represented as the anomaly map. 
	Through experiments using real-world datasets including PlantVillage, MVTec AD, and Cloud datasets, we demonstrated that the proposed layer-wise 
	external attention mechanism consistently boosts the anomaly detection performance of baseline models. We also confirmed that boosting abilities can remain 
	even for significantly imbalanced datasets. For intuitive understanding of the proposed attention mechanism, we visualized and compared feature maps output 
	from intermediate layers in Anomaly Detection Network~(ADN) before and after layer-wise external attention. 
	The results displayed that attention maps generated from Color Anomaly Attention Network (CAAN) successfully highlighted 
	anomalous regions. 
	
	In this paper, we experimentally demonstrated that our new attention mechanism considerably boosts the anomaly detection performance of CNN-based 
	image classifiers. We conclude that the layer-wise external attention is a promising technique for practical image anomaly detection. 
	Moreover, we would like to emphasize that the technique is applicable not only for image anomaly detection, but also for various image recognition tasks. 
	While, we found that the appropriate setting for successful boosting depends on the model and the dataset.
	Further studies focusing on theoretical aspects are expected to control adequately the layer-wise external attention.

\section{Acknowledgments}
	This work was supported by JST, PRESTO Grant Number JPMJPR1875, Japan. 

\begin{quote}
	\begin{small}
		\bibliography{reference}
	\end{small}
\end{quote}

\section{Additional Experiment for Anomaly Detection}
	We performed additional experiments to evaluate the performance of LEA-Net using Hurricane \cite{satellite}, Concrete \cite{concreate}, PlantVillage \cite{Hughes_2016}, MVTec \cite{Bergmann_2019}, Cloud \cite{cloud} datasets. 
	In the experiments, we additionally evaluate our layer-wise external attention on basic small CNN to see the effectiveness for various types of models.
	These experimental results were omitted from the main body of this paper due to the space constraints.

	\subsection{Additional Datasets}
\begin{figure}[htb]
			\centering
			\captionsetup{justification=raggedright, singlelinecheck=off, labelfont=bf}
			\includegraphics[width=0.45\textwidth]{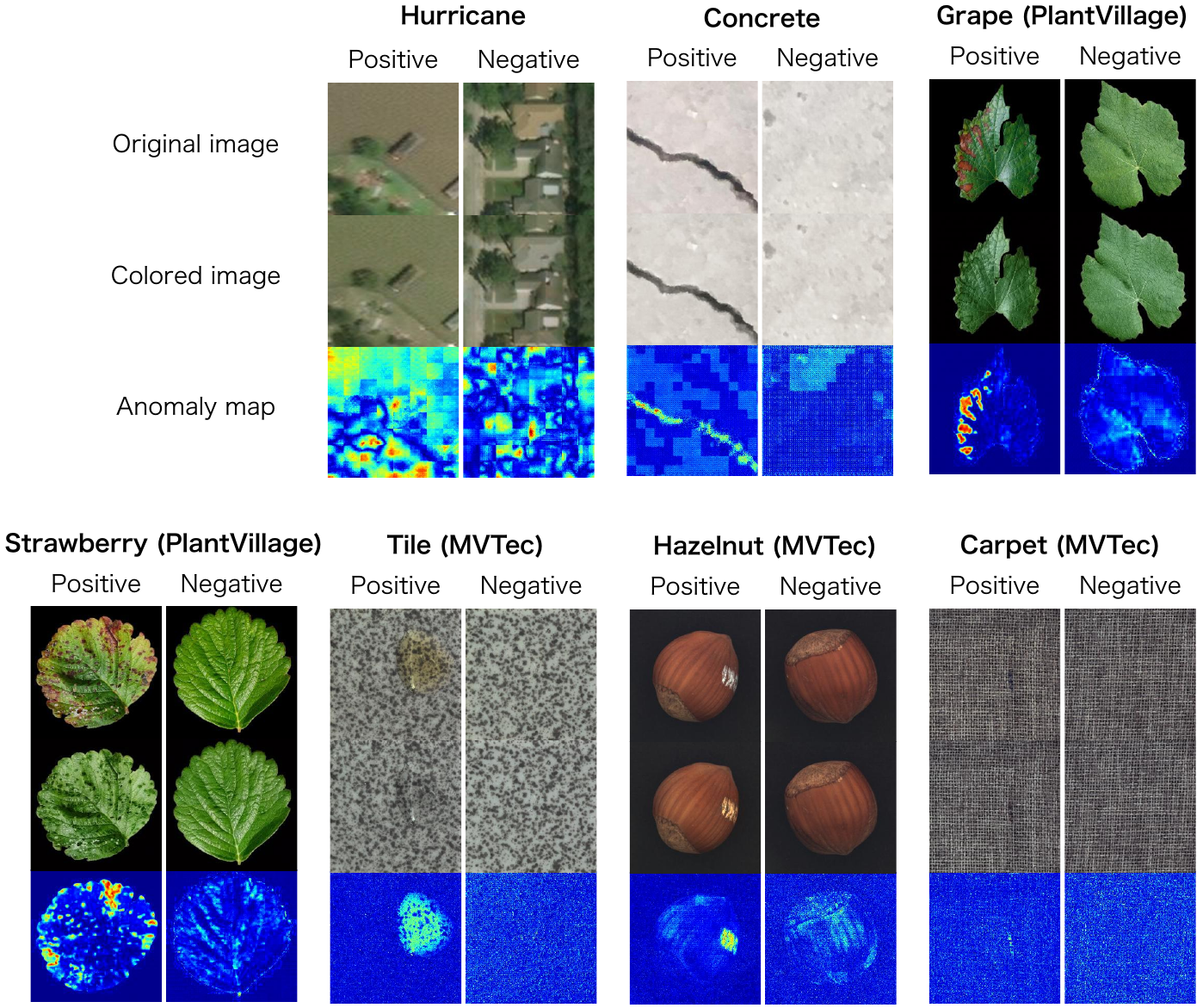}
			\caption{Examples of color anomaly maps for additional datasets. 
					For each dataset, the original and colored image and the anomaly map calculated from the two images are shown for Positive and Negative instances.
					Anomalous regions in the Positive images are failed to be colored, and the regions are highlighted in the color anomaly maps.}
			\label{fig:Add_Dataset}
		\end{figure}

		From PlantVillage, we used \textit{Grape} and \textit{Strawberry} datasets.
		From MVTec AD, we used \textit{Hazelnut}, \textit{Tile} and \textit{Carpet}, which contain color anomalous instances.
		Hurricane dataset contains normal (negative) images and anomalous (positive) images. The anomalous images show damages caused by hurricanes such as floods.
		Concrete dataset contains normal images without damage and anomalous images with cracks and splits.
		All images were resized to $256\times256$ pixels before experiments.
		The details of datasets are shown in Table~\ref{tab:Add_Dataset} and Fig.~\ref{fig:Add_Dataset}.

		In Fig.~\ref{fig:Add_Dataset}, we display illustrative examples of images and color anomaly maps for the dataset described above. 
		The first rows show original color images. 
		The second rows show reconstructed color images. 
		The third rows show anomaly maps. 
		Anomalous regions in the positive images are failed to be colored, and the regions are highlighted in the color anomaly maps.

	\subsection{Performance on Anomaly Detection}
		We show the comparison of anomaly detection performance between LEA-Net and other networks in Table~\ref{tab:Add_Performance_1}-\ref{tab:Add_Performance_4}.
		The performances are evaluated on the averages of $F_1$ scores with standard deviation.
		The best and second-best performances are emphasized by bold type.
		The values given in parentheses indicate attention point of the best $F_1$ score.

		Table~\ref{tab:Add_Performance_1} and Table~\ref{tab:Add_Performance_2} show that 
		LEA-Nets (networks with MobileNet-small or ResNet-based CAAN) demonstrated the best or second-best performances in most of the cases.
		From these results, it is shown that the layer-wise external attention mechanism consistently boosted the anomaly detection performances of ResNet18 and VGG16. 
		In addition, Table~\ref{tab:Add_Performance_3} and Table~\ref{tab:Add_Performance_4} clearly show that the layer-wise external attention mechanism works well for basic CNN.
		The $F_1$ scores of basic CNN based LEA-Net are the hignest or second highest for most of datasets.
		From this additional experimental results, we confirmed that the layer-wise external attention mechanism is applicable and effective for various types of models.
		
		\begin{table}[h]
			\centering
			\captionsetup{labelfont=bf}
			\caption{Additional experimental datasets.}
			\label{tab:Add_Dataset}
			\scalebox{0.8}[0.8]{ 
			\renewcommand{\arraystretch}{1.2}
			\begin{tabular}{c|c c c c}
				
				Dataset Name
				& Positive 
				& Negative 
				& \begin{tabular}{c}
					P to N \\[-0.5em] Ratio
				\end{tabular} 
				& Total \\ \hline
				
				\textit{Hurricane}  & 100 & 300 & 0.333 & 400 \\
				\textit{Concrete}   & 100 & 300 & 0.333 & 400 \\
				\textit{Grape}      & 141 & 423 & 0.333 & 564 \\
				\textit{Strawberry} & 152 & 456 & 0.333 & 608 \\
				\textit{Tile}       & 84  & 263 & 0.319 & 347  \\
				\textit{Hazelnut}   & 70  & 431 & 0.162 & 501  \\
				\textit{Carpet}     & 89  & 308 & 0.289 & 397  \\

			\end{tabular}
			\renewcommand{\arraystretch}{1}
			}
		\end{table}

		\begin{table*}[t]
			\centering
			\captionsetup{labelfont=bf}
			\caption{Comparison of $F_1$ scores on \textit{Hurricane}, \textit{Concrete}, \textit{Grape} and \textit{Strawberry} datasets.}
			\label{tab:Add_Performance_1}
			\scalebox{0.9}[0.9]{ 
			\renewcommand{\arraystretch}{1.5}
			\begin{tabular}{l|c||c|c|c|c}
				
				Model      &  
				Params     & 
				Hurricane  & 
				Concrete   &
				Grape      &
				Strawberry \\ 
				
				\hline 
				
				\small{ResNet18}           	
									& 11.2M              	
									& \textbf{0.874 $\pm$ 0.075}
									& \textbf{0.974 $\pm$ 0.026}
									& 0.944 $\pm$ 0.044
									& 0.840 $\pm$ 0.197  \\

				\small{ResNet50}
									& 23.6M 
									& 0.781 $\pm$ 0.125
									& 0.921 $\pm$ 0.026
									& 0.892 $\pm$ 0.062
									& 0.820 $\pm$ 0.163  \\

				\small{ResNet18 (AnomalyMap input)}
									& 11.2M 
									& 0.403 $\pm$ 0.007
									& 0.790 $\pm$ 0.218
									& 0.856 $\pm$ 0.133
									& 0.778 $\pm$ 0.226  \\

				\small{ResNet18 (4ch input)}
									& 11.2M 
									& 0.412 $\pm$ 0.013
									& 0.946 $\pm$ 0.035
									& 0.918 $\pm$ 0.027
									& 0.866 $\pm$ 0.137  \\

				\small{ResNet18 (Attentioned input)}
									& 11.2M 
									& 0.500 $\pm$ 0.076
									& 0.952 $\pm$ 0.029
									& 0.895 $\pm$ 0.101
									& 0.899 $\pm$ 0.095  \\
				
				\small{ResNet18 + Direct Attention}
									& 11.2M 
									& \textbf{0.870 $\pm$ 0.058(5)}
									& 0.963 $\pm$ 0.036(5)
									& 0.951 $\pm$ 0.046(1)
									& \textbf{0.986 $\pm$ 0.019(2)}  \\

				\small{\textbf{ResNet18 + MobileNet-small CAAN}}
									& 13.5M 
									& 0.858 $\pm$ 0.076(3)
									& \textbf{0.979 $\pm$ 0.029(1)}
									& \textbf{0.978 $\pm$ 0.039(4)}
									& \textbf{0.987 $\pm$ 0.022(4)}  \\

				\small{\textbf{ResNet18 + ResNet-based CAAN}}
									& 16.1M 
									& 0.866 $\pm$ 0.054(5)
									& 0.969 $\pm$ 0.022(3)
									& \textbf{0.961 $\pm$ 0.048(2)}
									& 0.957 $\pm$ 0.054(3)  \\

				\hline

				\small{VGG16}
									& 165.7M              	
									& 0.702 $\pm$ 0.092
									& 0.959 $\pm$ 0.029
									& 0.982 $\pm$ 0.026
									& 0.959 $\pm$ 0.033  \\

				\small{VGG19}
									& 171.0M 
									& 0.655 $\pm$ 0.065
									& \textbf{0.980 $\pm$ 0.011}
									& 0.971 $\pm$ 0.016
									& 0.963 $\pm$ 0.03  \\

				\small{VGG16 (AnomalyMap input)}     
									& 165.7M 
									& 0.400 $\pm$ 0.000
									& 0.886 $\pm$ 0.028
									& 0.902 $\pm$ 0.051
									& 0.849 $\pm$ 0.109  \\

				\small{VGG16 (4ch input)}
									& 165.7M 
									& 0.441 $\pm$ 0.034
									& \textbf{0.942 $\pm$ 0.036}
									& 0.956 $\pm$ 0.037
									& 0.765 $\pm$ 0.158  \\

				\small{VGG16 (Attentioned input)}        	
									& 165.7M 
									& 0.584 $\pm$ 0.119
									& 0.938 $\pm$ 0.047
									& 0.978 $\pm$ 0.015
									& 0.966 $\pm$ 0.025  \\
				
				\small{VGG16 + Direct Attention}
									& 165.7M 
									& \textbf{0.717 $\pm$ 0.029(5)}
									& 0.969 $\pm$ 0.021(5)
									& 0.982 $\pm$ 0.013(3)
									& 0.968 $\pm$ 0.039(1)  \\

				\small{\textbf{VGG16 + MobileNet-small CAAN}}
									& 168.0M 
									& 0.715 $\pm$ 0.084(5)
									& 0.964 $\pm$ 0.023(1)
									& \textbf{0.986 $\pm$ 0.015(1)}
									& \textbf{0.969 $\pm$ 0.027(3)}  \\

				\small{\textbf{VGG16 + ResNet-based CAAN}}
									& 170.6M 
									& \textbf{0.725 $\pm$ 0.045(4)}
									& 0.964 $\pm$ 0.023(3)
									& \textbf{0.986 $\pm$ 0.015(1)}
									& \textbf{0.973 $\pm$ 0.020(2)}  \\
		
			\end{tabular}
			\renewcommand{\arraystretch}{1}
			}
		\end{table*}
		
		\begin{table*}[t]
			\centering
			\captionsetup{labelfont=bf}
			\caption{Comparison of $F_1$ scores on \textit{Tile}, \textit{Hazelnut} and \textit{Carpet} datasets.}
			\label{tab:Add_Performance_2}
			\scalebox{0.9}[0.9]{ 
			\renewcommand{\arraystretch}{1.5}
			\begin{tabular}{l|c||c|c|c}
				
				Model      &  
				Params     & 
				Tile       &   
				Hazelnut   & 
				Carpet     \\ 
				
				\hline 
				
				\small{ResNet18}           	
					& 11.2M              	
					& 0.768 $\pm$ 0.120
					& 0.866 $\pm$ 0.021
					& 0.711 $\pm$ 0.061  \\

				\small{ResNet50}
					& 23.6M 
					& 0.601 $\pm$ 0.228 
					& 0.779 $\pm$ 0.302
					& 0.524 $\pm$ 0.151  \\
								
				\small{ResNet18 (AnomalyMap input)}
					& 11.2M 
					& 0.381 $\pm$ 0.103 
					& 0.635 $\pm$ 0.06
					& 0.392 $\pm$ 0.195  \\

				\small{ResNet18 (4ch input)}
					& 11.2M
					& 0.669 $\pm$ 0.190
					& 0.854 $\pm$ 0.193
					& 0.686 $\pm$ 0.06  \\

				\small{ResNet18 (Attentioned input)}
					& 11.2M 
					& 0.750 $\pm$ 0.136 
					& 0.787 $\pm$ 0.304
					& 0.735 $\pm$ 0.057  \\
									
				\small{ResNet18 + Direct Attention}
					& 11.2M 
					& \textbf{0.780 $\pm$ 0.077(5)}
					& \textbf{0.902 $\pm$ 0.075(1)}
					& \textbf{0.756 $\pm$ 0.045(3)}  \\

				\small{\textbf{ResNet18 + MobileNet-small CAAN}}
					& 13.5M 
					& \textbf{0.789 $\pm$ 0.096(2)} 
					& \textbf{0.919 $\pm$ 0.079(2)}
					& \textbf{0.788 $\pm$ 0.057(3)}  \\
								
				\small{\textbf{ResNet18 + ResNet-based CAAN}}
					& 16.1M 
					& 0.776 $\pm$ 0.103(3)
					& 0.897 $\pm$ 0.040(5)
					& 0.724 $\pm$ 0.055(1)  \\

				\hline

				\small{VGG16}
					& 165.7M              	
					& 0.346 $\pm$ 0.094
					& 0.857 $\pm$ 0.082
					& 0.373 $\pm$ 0.349  \\

				\small{VGG19}
					& 171.0M 
					& \textbf{0.835 $\pm$ 0.091} 
					& \textbf{0.903 $\pm$ 0.057}
					& \textbf{0.634 $\pm$ 0.304}  \\
									
				\small{VGG16 (AnomalyMap input)}     
					& 165.7M
					& \textbf{0.599 $\pm$ 0.115} 
					& 0.7 $\pm$ 0.092
					& 0.423 $\pm$ 0.177  \\ 

				\small{VGG16 (4ch input)}
					& 165.7M 	
					& 0.589 $\pm$ 0.103
					& 0.78 $\pm$ 0.129
					& \textbf{0.593 $\pm$ 0.137}  \\

				\small{VGG16 (Attentioned input)}        	
					& 165.7M 
					& 0.512 $\pm$ 0.134
					& 0.791 $\pm$ 0.074
					& 0.252 $\pm$ 0.141  \\
									
				\small{VGG16 + Direct Attention}
					& 165.7M 
					& 0.399 $\pm$ 0.103(2) 
					& \textbf{0.842 $\pm$ 0.068(3)}
					& 0.479 $\pm$ 0.108(3)  \\
								
				\small{\textbf{VGG16 + MobileNet-small CAAN}}
					& 168.0M 
					& 0.432 $\pm$ 0.177(2)
					& 0.828 $\pm$ 0.066(4)
					& 0.55 $\pm$ 0.186(4)  \\

				\small{\textbf{VGG16 + ResNet-based CAAN}}
					& 170.6M 
					& 0.525 $\pm$ 0.128(1) 
					& 0.82 $\pm$ 0.057(1)
					& 0.484 $\pm$ 0.247(4)  \\
									
			\end{tabular}
			\renewcommand{\arraystretch}{1}
			}
		\end{table*}
		
		\begin{table*}[t]
			\captionsetup{labelfont=bf}
			\caption{Comparison of $F_1$ scores on \textit{Retina}, \textit{Potato}, \textit{Cloud} and \textit{Leather} datasets.}
			\label{tab:Add_Performance_3}
			\scalebox{0.9}[0.9]{ 
			\renewcommand{\arraystretch}{1.5}
			\begin{tabular}{l|c||c|c|c|c}
				
				Model   &  
				Params  & 
				Retina  &
				Potato  & 
				Cloud   & 
				Leather \\
				
				\hline 					
				\small{basic CNN}           	
									& 4.0M              	
									& 0.621 $\pm$ 0.027
									& 0.868 $\pm$ 0.066
									& 0.276 $\pm$ 0.131
									& 0.54 $\pm$ 0.094 \\ 	

				\small{basic CNN (AnomalyMap input)}
									& 4.0M 
									& 0.417 $\pm$ 0.036
									& 0.772 $\pm$ 0.018
									& \textbf{0.475 $\pm$ 0.056}
									& 0.45 $\pm$ 0.191\\

				\small{basic CNN (4ch input)}
									& 4.0M  
									& 0.483 $\pm$ 0.052
									& 0.829 $\pm$ 0.064
									& \textbf{0.421 $\pm$ 0.057}
									& \textbf{0.613 $\pm$ 0.111} \\

				\small{basic CNN (Attentioned input)}
									& 4.0M  
									& 0.63 $\pm$ 0.049
									& 0.858 $\pm$ 0.058
									& 0.383 $\pm$ 0.109
									& 0.459 $\pm$ 0.087 \\
				
				\small{basic CNN + Direct Attention}
									& 4.0M 
									& 0.65 $\pm$ 0.055(3)
									& 0.884 $\pm$ 0.051(5)
									& 0.332 $\pm$ 0.103(3)
									& \textbf{0.625 $\pm$ 0.11(5)}\\ 

				\small{\textbf{basic CNN + MobileNet-small CAAN}}
									& 7.0M 
									& \textbf{0.652 $\pm$ 0.015(5)}
									& \textbf{0.925 $\pm$ 0.033(5)}
									& 0.324 $\pm$ 0.09(4)
									& 0.606 $\pm$ 0.093(3)\\ 

				\small{\textbf{basic CNN + ResNet-based CAAN}}
									& 8.9M 
									& \textbf{0.654 $\pm$ 0.052(3)}
									& \textbf{0.907 $\pm$ 0.074(4)}
									& 0.337 $\pm$ 0.111(4)
									& 0.591 $\pm$ 0.134(4) \\ 
			\end{tabular}
			\renewcommand{\arraystretch}{1}
			}
		\end{table*}
		
		\begin{table*}[t]
			\centering
			\captionsetup{labelfont=bf}
			\caption{Comparison of $F_1$ scores on \textit{Hurricane}, \textit{Concrete}, \textit{Grape} and \textit{Hazelnut} datasets.}
			\label{tab:Add_Performance_4}
			\scalebox{0.9}[0.9]{ 
			\renewcommand{\arraystretch}{1.5}
			\begin{tabular}{l|c||c|c|c|c}
				
				Model      &  
				Params     & 
				Hurricane  & 
				Concrete   &
				Grape      &
				Hazelnut    \\ 
				
				\hline 
				
				\small{basic CNN}           	
									& 4.0M              	
									& 0.835 $\pm$ 0.072
									& 0.875 $\pm$ 0.071
									& 0.955 $\pm$ 0.017
									& 0.463 $\pm$ 0.172  \\

				\small{basic CNN (AnomalyMap input)}
									& 4.0M 
									& 0.401 $\pm$ 0.002
									& 0.795 $\pm$ 0.109
									& 0.868 $\pm$ 0.1
									& 0.503 $\pm$ 0.183\\

				\small{basic CNN (4ch input)}
									& 4.0M 
									& 0.44 $\pm$ 0.024
									& 0.888 $\pm$ 0.055
									& 0.917 $\pm$ 0.074
									& \textbf{0.683 $\pm$ 0.137}\\

				\small{basic CNN (Attentioned input)}
									& 4.0M 
									& 0.601 $\pm$ 0.18
									& 0.882 $\pm$ 0.057
									& 0.941 $\pm$ 0.03
									& 0.377 $\pm$ 0.231\\
				
				\small{basic CNN + Direct Attention}
									& 4.0M 
									& \textbf{0.848 $\pm$ 0.072(3)}
									& 0.905 $\pm$ 0.049(3)
									& \textbf{0.96 $\pm$ 0.024(3)}
									& 0.465 $\pm$ 0.151(4)\\

				\small{\textbf{basic CNN + MobileNet-small CAAN}}
									& 7.0M 
									& 0.847 $\pm$ 0.089(4)
									& \textbf{0.918 $\pm$ 0.062(5)}
									& 0.956 $\pm$ 0.017(2)
									& 0.462 $\pm$ 0.166(4)\\

				\small{\textbf{basic CNN + ResNet-based CAAN}}
									& 8.9M 
									& \textbf{0.863 $\pm$ 0.101(2)}
									& \textbf{0.924 $\pm$ 0.051(5)}
									& \textbf{0.959 $\pm$ 0.029(4)}
									& \textbf{0.545 $\pm$ 0.261(5)}\\
		
			\end{tabular}
			\renewcommand{\arraystretch}{1}
			}
		\end{table*}


\newpage
\section{U-Net Architecture}

	In Table ~\ref{tab:Unet}, we describe the architecture of U-Net that was used in the color anomaly map generation step.
	The sizes of input images are $256 \times 256$ pixels, and there are eight downsampling points.
	Except for the first and the last convolutional and deconvolutional layers, batch normalization layers are followed by the convolutional and the deconvolutional layers.

	\begin{table*}[ht]
		\centering
		\captionsetup{labelfont=bf}
		\caption{The U-Net architecture used in the present study. Conv2D denotes a 2D convolutional layer and BN denotes a batch normalization. All alpha values of LeakyRelu are $0.2$.}
		\renewcommand{\arraystretch}{1.2}
		\label{tab:Unet}
		\begin{tabular}{r r|c c c c c}
			
			ID   &  
			Layer Type  & 
			Output dim  &
			Kernel & 
			Stride   & 
			Filters & 
			Activation    \\ 
			
			\hline 

			1 &
			Input &
			256 $\times$ 256 &
			- &
			- &
			- &
			- \\

			2 &
			Conv2D &
			128 $\times$ 128 &
			4 &
			2 &
			64 &
			LeakyRelu \\

			3 &
			Conv2D $\rightarrow$ BN &
			64 $\times$ 64 &
			4 &
			2 &
			128 &
			LeakyRelu \\

			4 &
			Conv2D $\rightarrow$ BN &
			32 $\times$ 32 &
			4 &
			2 &
			256 &
			LeakyRelu \\

			5 &
			Conv2D $\rightarrow$ BN &
			16 $\times$ 16 &
			4 &
			2 &
			512 &
			LeakyRelu \\

			6 &
			Conv2D $\rightarrow$ BN &
			8 $\times$ 8 &
			4 &
			2 &
			512 &
			LeakyRelu \\

			7 &
			Conv2D $\rightarrow$ BN &
			4 $\times$ 4 &
			4 &
			2 &
			512 &
			LeakyRelu \\

			8 &
			Conv2D $\rightarrow$ BN &
			2 $\times$ 2 &
			4 &
			2 &
			512 &
			LeakyRelu \\

			9 &
			Conv2D $\rightarrow$ BN &
			1 $\times$ 1 &
			4 &
			2 &
			512 &
			LeakyRelu \\

			10 &
			Conv2DTranspose$\rightarrow$ BN &
			2 $\times$ 2 &
			2 &
			2 &
			512 &
			Relu \\

			11 & 
			Concat(8, 10) & 
			2 $\times$ 2 & 
			- &
			- &
			1024 &
			- \\

			12 &
			Conv2DTranspose$\rightarrow$ BN &
			4 $\times$ 4 &
			2 &
			2 &
			512 &
			Relu \\

			13 & 
			Concat(7, 12) & 
			4 $\times$ 4 & 
			- &
			- &
			1024 &
			- \\

			14 &
			Conv2DTranspose$\rightarrow$ BN &
			8 $\times$ 8 &
			2 &
			2 &
			512 &
			Relu \\

			15 & 
			Concat(6, 14) & 
			8 $\times$ 8 & 
			- &
			- &
			1024 &
			- \\

			16 &
			Conv2DTranspose$\rightarrow$ BN &
			16 $\times$ 16 &
			2 &
			2 &
			512 &
			Relu \\

			17 & 
			Concat(5, 16) & 
			16 $\times$ 16 & 
			- &
			- &
			1024 &
			- \\

			18 &
			Conv2DTranspose$\rightarrow$ BN &
			32 $\times$ 32 &
			2 &
			2 &
			256 &
			Relu \\

			19 & 
			Concat(4, 18) & 
			32 $\times$ 32 & 
			- &
			- &
			512 &
			- \\

			20 &
			Conv2DTranspose$\rightarrow$ BN &
			64 $\times$ 64 &
			2 &
			2 &
			128 &
			Relu \\

			21 & 
			Concat(3, 20) & 
			64 $\times$ 64 & 
			- &
			- &
			256 &
			- \\

			22 &
			Conv2DTranspose$\rightarrow$ BN &
			128 $\times$ 128 &
			2 &
			2 &
			64 & 
			Relu \\

			23 & 
			Concat(2, 22) & 
			128 $\times$ 128 & 
			- &
			- &
			128 &
			- \\

			24 &
			Conv2DTranspose &
			256 $\times$ 256 &
			2 &
			2 &
			2 &
			sigmoid \\

		\end{tabular}
		\renewcommand{\arraystretch}{1}
	\end{table*}


\end{document}